\documentclass[11pt]{article}

\usepackage[final]{acl}
 \usepackage{microtype}

\usepackage{graphicx}
\usepackage{booktabs}
\usepackage{amsmath}
\usepackage{amssymb}
\usepackage{multirow}
 \usepackage{times}
\usepackage{latexsym}
\usepackage[T1]{fontenc}
\usepackage[utf8]{inputenc}

\usepackage[dvipsnames]{xcolor}
\everydisplay{\footnotesize}
\usepackage{tikz}
\usetikzlibrary{positioning,calc,fit,arrows.meta,backgrounds}

\usepackage[english,bidi=default]{babel} % English as the main language.
\babelfont{rm}{TeXGyreTermesX} % similar to Times
\babelprovide[import]{hindi}
\babelfont[*devanagari]{rm}{Lohit Devanagari}
\babelprovide[import]{arabic}

\definecolor{maskgrey}{RGB}{226,226,230}
\definecolor{maskline}{RGB}{138,138,146}
\definecolor{masktext}{RGB}{92,92,100}
\definecolor{keepblue}{RGB}{204,223,244}
\definecolor{keepline}{RGB}{45,105,175}
\definecolor{keeptext}{RGB}{22,64,120}
\definecolor{tmplbg}{RGB}{253,235,211}
\definecolor{tmplline}{RGB}{197,133,32}
\definecolor{tem}{RGB}{203,238,255}
\definecolor{mypink}{RGB}{255, 222, 230}

\title{alkhder at AlexandriaX-2026: Context-Aware English-to-Dialectal Arabic Dialogue Translation with QLoRA-Fine-Tuned NileChat-3B}
\author{
  Hasan Alkhder \\
  Sakarya University, T\"urkiye \\
  \small\texttt{hasan.alkhder2@ogr.sakarya.edu.tr}
  \And
  Maher Pasa \\
  Altınbaş University, T\"urkiye \\
  \small\texttt{engmaherpasa@gmail.com}
  \And
  Maria Hanini \\
  Imperial College London \\
  \small\texttt{m.hanini@imperial.ac.uk}
  \AND
  Amro Najjar \\
  Luxembourg Institute of Science and Technology \\
  \small\texttt{amro.najjar@list.lu}
  \And
  Ahmet Zengin \\
  Sakarya University, T\"urkiye \\
  \small\texttt{azengin@sakarya.edu.tr}
}
\begin{document}

\maketitle
\begin{abstract}

This paper describes team Alkhder's submission to Subtask~1 of the AlexandriaX-2026 shared task, which asks systems to translate English dialogue turns into thirteen Arabic dialects. We fine-tune NileChat-3B-Base with QLoRA on the Alexandria parallel dialogue corpus, training one set of adapter weights for all dialects. Each turn is translated conditioned on the target country and dialect, the domain, the participant roles, the speaker–addressee gender direction, and the preceding dialogue context. At inference, conversations are translated autoregressively and the model is fed its own previous outputs back as context without using any gold target-side test information. Under the constrained track, using only Alexandria data, it trains in about one hour and translates the private test set in under two hours. The system reaches a country-macro-averaged spBLEU of 20.36 and chrF++ of 36.64 on both tracks.
\end{abstract}

 \section{Introduction}

Dialectal Arabic machine translation (MT) remains difficult because Arabic varieties differ from Modern Standard Arabic (MSA) and from one another in vocabulary, morphology, syntax, pragmatics, and sociolinguistic use, including across cities within the same country~\citep{abdul2024survey}. Progress is limited by scarce and domain-specific parallel data~\citep{meftouh2015machine, bouamor2018madar,mohamed2023alphamwe}, non-standardized spelling ~\citep{alhafni2024exploiting, kheder2024aladan}, and the need to use dialogue context and speaker information during translation~\citep{elmekki-etal-2026-alexandriax, bawden-etal-2016-investigating}.  The AlexandriaX-2026 shared task~\citep{elmekki-etal-2026-alexandriax} addresses these issues by evaluating systems on meaning preservation and dialect-specific adaptation.

We participate in Subtask~1, \emph{Context-Aware English-to-Dialectal Arabic Dialogue Translation}, in both the constrained and unconstrained track. The task is to translate selected English dialogue turns into one of thirteen dialectal Arabic varieties, using the available conversational and metadata context. Two varieties, Libyan and Sudanese, have no training data and are therefore evaluated in a zero-shot setting.

We build one system that satisfies the constrained track rules and submit it to both the constrained and unconstrained rankings. The system fine-tunes NileChat-3B-Base~\citep{el-mekki-etal-2025-nilechat}, a 3B-parameter Arabic and dialect-focused language model, with QLoRA~\citep{dettmers2023qlora} on turn-level prompts derived from the Alexandria corpus~\citep{el-mekki-etal-2026-alexandria}. Each prompt includes the target dialect and the relevant dialogue metadata. One set of adapter weights is used for all dialects.\footnote{Code: \url{https://github.com/HasanAlkheder/AlexandriaX2026-Subtask1}.}

We show that this compact approach trains in about one GPU-hour and covers all dialects. The system achieves a country-macro-averaged spBLEU of 20.36 and chrF++ of 36.64, ranking sixth in both tracks. We show that, as implemented, self-generated dialogue context contributes no measurable benefit. We further discuss the challenge imposed by the strong per-country resource imbalance in the training data against the country-macro-averaged evaluation metric.

The paper proceeds as follows. Section~\ref{sec:background} covers the task setup and related work, Section~\ref{sec:system} our system, and Section~\ref{sec:setup} the experimental configuration. Section~\ref{sec:results} reports results, before concluding in Section~\ref{sec:conclusion}.

\section{Background}
\label{sec:background}

\subsection{Task setup}
\label{sec:taskdesc}

AlexandriaX-2026~\citep{elmekki-etal-2026-alexandriax} includes three subtasks: context-aware English-to-dialectal Arabic dialogue translation, cross-dialect Arabic MT for short financial queries, and diagnostic MT evaluation based on error spans and an LQM-inspired typology~\citep{malaysha-etal-2024-arafinnlp,magdy-etal-2026-lqm}. We participate in Subtask~1, which requires systems to translate selected English turns into a specified Arabic dialect using dialogue context, country and dialect labels, domain, persona descriptions, speaker--addressee gender direction and turn position. The translation must preserve meaning and use appropriate lexical, morphological, pragmatic and sociolinguistic forms. We enter both tracks. The constrained track allows only Alexandria data and models with up to 5B parameters, whereas the unconstrained track permits external data and models of any size. 

\subsection{Data}

The task uses the Alexandria corpus~\citep{el-mekki-etal-2026-alexandria}, which contains multi-domain, multi-turn English--dialectal Arabic conversations from thirteen countries. Training and development data are available for eleven countries and contain 66{,}480 and 12{,}250 turns, respectively. The data are highly imbalanced, with 14{,}933 Palestinian training turns compared with 2{,}034 Tunisian turns (Table~\ref{tab:data} reports the train+dev pool). Libya and Sudan have no training or development data and are evaluated zero-shot. The public test set contains 14{,}442 turns, while the private test set contains 14{,}459 turns across 4{,}673 conversations, distributed near-uniformly over the thirteen countries (roughly 1{,}100 turns each, with 1{,}309 for Libya and 915 for Sudan) and scored through CodaBench.

\begin{table}[htbp!]
\centering
\footnotesize
\begin{tabular}{lrr}
\toprule
\textbf{Country} & \textbf{Available} & \textbf{Used (cap 2{,}500)} \\
\midrule
EG (Egypt)        & 4{,}221  & 2{,}500 \\
JO (Jordan)       & 6{,}614  & 2{,}500 \\
LB (Lebanon)      & 10{,}024 & 2{,}500 \\
MA (Morocco)      & 3{,}683  & 2{,}500 \\
MR (Mauritania)   & 6{,}629  & 2{,}500 \\
OM (Oman)         & 7{,}389  & 2{,}500 \\
PS (Palestine)    & 16{,}043 & 2{,}500 \\
SA (Saudi Arabia) & 9{,}580  & 2{,}500 \\
SY (Syria)        & 7{,}190  & 2{,}500 \\
TN (Tunisia)      & 3{,}150  & 2{,}500 \\
YE (Yemen)        & 4{,}207  & 2{,}500 \\
\midrule
Total             &78{,}730 & 27{,}500 \\
\bottomrule
\end{tabular}
\caption{Turn-level training examples per country. Libya and Sudan have no training data.}
\label{tab:data}
\end{table}
\subsection{Related Work}
\label{sec:relwork}

Early dialectal Arabic MT addressed limited data through crowdsourcing and MSA-based pivoting~\citep{zbib2012machine,salloum-habash-2013-dialectal}. Later resources such as PADIC, MADAR and AraBench expanded corpus coverage and evaluation, but remained limited by restricted domains and data scarcity~\citep{meftouh2015machine,bouamor2018madar,sajjad-etal-2020-arabench}. Arabic-specific pretrained models and instruction-tuned LLMs have improved performance, although results remain uneven across dialects~\citep{kadaoui-etal-2023-tarjamat,el-mekki-etal-2025-nilechat,alkhder2026sdnlp,jon-etal-2026-current}.

Arabic dialects also lack standardized orthographies, which creates multiple valid spellings and affects automatic evaluation. Existing approaches normalize dialectal text into CODA or MSA~\citep{habash-etal-2018-unified,alhafni2024exploiting,khered2025dial2msa}. Context-aware MT further shows that previous dialogue turns improve coherence, pronoun resolution and ambiguity resolution, while speaker information such as gender helps reduce agreement errors~\citep{tiedemann2017neuralmachinetranslationextended,voita-etal-2018-context,bawden-etal-2018-evaluating,vanmassenhove-etal-2018-getting}. AlexandriaX-2026 extends this work to English-to-dialectal Arabic translation across thirteen varieties by providing dialogue history and speaker--addressee gender metadata.

Recent low-resource MT systems adapt pretrained multilingual models~\citep{degenaro-lupicki-2024-experiments}. Parameter-efficient methods such as LoRA and QLoRA make this feasible with limited computational resources \citep{hu2022lora,dettmers2023qlora}.

 \section{System Overview}
\label{sec:system}
 
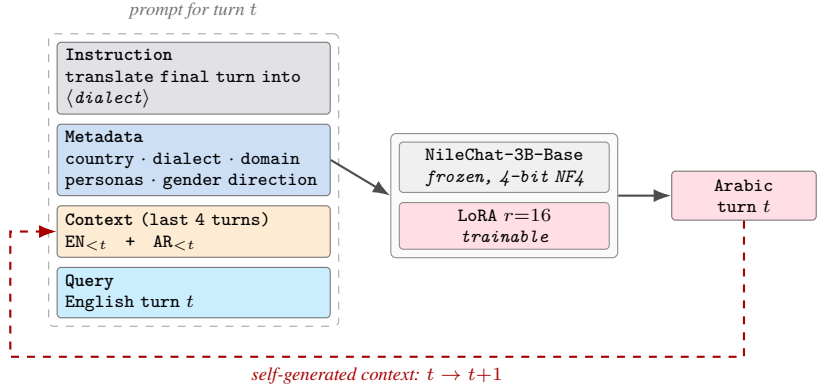
\begin{figure*}[t]
\centering
\begin{tikzpicture}[
  font=\scriptsize,
  >=Latex,
  block/.style={draw=black!55, rounded corners=1.5pt, align=left,
                inner xsep=3pt, inner ysep=2.5pt, text width=34mm, font=\scriptsize\ttfamily},
  instr/.style={block, fill=maskgrey, font=\scriptsize\ttfamily},
  meta/.style={block, fill=keepblue, font=\scriptsize\ttfamily},
  ctx/.style={block, fill=tmplbg, font=\scriptsize\ttfamily},
  query/.style={block, fill=tem, font=\scriptsize\ttfamily},
  sub/.style={draw=black!45, rounded corners=1.5pt, align=center,
              inner sep=2.5pt, text width=26mm, font=\scriptsize\ttfamily},
  frozen/.style={sub, fill=black!6, font=\scriptsize\ttfamily},
  train/.style={sub, fill=mypink, font=\scriptsize\ttfamily},
  outbox/.style={draw=black!55, rounded corners=1.5pt, fill=mypink,
                 align=center, inner sep=3pt, text width=17mm, font=\scriptsize\ttfamily},
  arr/.style={->, thick, black!70},
  fb/.style={->, thick, dashed, red!65!black},
  tag/.style={font=\scriptsize\itshape, text=black!55}
]

% ---- prompt blocks ----
\node[instr] (instr) {\textbf{Instruction}\\ translate final turn into
                      \textit{$\langle$dialect$\rangle$}};
\node[meta, below=1.2mm of instr] (meta)
  {\textbf{Metadata}\\ country $\cdot$ dialect $\cdot$ domain\\
   personas $\cdot$ gender direction};
\node[ctx, below=1.2mm of meta] (ctx)
  {\textbf{Context} (last 4 turns)\\ EN$_{<t}$ \; + \; AR$_{<t}$};
\node[query, below=1.2mm of ctx] (query)
  {\textbf{Query}\\ English turn $t$};

\begin{scope}[on background layer]
  \node[draw=black!35, dashed, rounded corners=2pt, fit=(instr)(query),
        inner sep=3pt] (prompt) {};
\end{scope}
\node[tag, above=0.5mm of prompt.north] {prompt for turn $t$};

% ---- model ----
\node[frozen, right=9mm of ctx.east, anchor=west, yshift=8.5mm] (base)
  {NileChat-3B-Base\\ \textit{frozen, 4-bit NF4}};
\node[train, below=1.2mm of base] (lora)
  {LoRA $r{=}16$\\ \textit{trainable}};
\begin{scope}[on background layer]
  \node[draw=black!45, rounded corners=2pt, fill=black!2,
        fit=(base)(lora), inner sep=3pt] (model) {};
\end{scope}

% ---- output ----
\node[outbox, right=7mm of model] (outp) {Arabic\\ turn $t$};

% ---- edges ----
\draw[arr] (meta.east) -- (model.west);
\draw[arr] (model.east) -- (outp.west);

% ---- feedback ----
\coordinate (down) at ($(outp.south)+(0,-18mm)$);
\coordinate (left) at ($(prompt.west)+(-5mm,0)$);
\draw[fb] (outp.south) -- (down) -- (down -| left)
          -- (ctx.west -| left) -- (ctx.west);
\node[tag, text=red!65!black, anchor=north, align=center]
  at ($(down)!0.5!(down -| left)$) {self-generated context: $t \rightarrow t{+}1$};

\end{tikzpicture}
\caption{System overview. }
\label{fig:pipeline}
\end{figure*}
\subsection{Prompt construction}
\label{sec:prompt}

Every training and test example is formatted as one instruction-style prompt with four blocks: (i) an \textbf{instruction} to translate only the final English turn into the named dialect, preserving meaning, tone, register, persona and gender agreement; (ii) a \textbf{metadata} block giving the target country and dialect name, domain, participant descriptions, and the current speaker with the gender direction (e.g.\ \textit{female $\rightarrow$ male}); (iii) the \textbf{context}, i.e.\ the conversation so far truncated to the last four turns, where each previous turn appears as its English source and, when available, its Arabic translation; and (iv) the \textbf{query}, the English turn to translate, followed by a fixed response marker. Appendix~\ref{app:prompt} provides a prompt template. 

The dialect is specified in natural language, allowing one adapter to serve all thirteen varieties, including the zero-shot Libyan and Sudanese dialects. Although Libyan and Sudanese are absent from fine-tuning, the model can associate them with related varieties seen during pretraining, namely Maghrebi Arabic for Libyan and Egyptian Arabic for Sudanese. Training prompts use gold Arabic translations in the context, whereas inference prompts use the model's own previous outputs.

\subsection{Base model and fine-tuning}
\label{sec:model}

As our base model, we select NileChat-3B-Base~\citep{el-mekki-etal-2025-nilechat}, a dialect-focused Arabic language model with $\approx$3B parameters leaving headroom under the 5B constrained track limit. We fine-tune with QLoRA~\citep{dettmers2023qlora}. The base weights are frozen and quantised to 4-bit NF4 with double quantisation and bfloat16 compute. Low-rank adapters~\citep{hu2022lora} with $r=16$, $\alpha=32$ and dropout $0.05$ are trained on all attention and MLP projection matrices (\texttt{q/k/v/o\_proj}, \texttt{gate/up/down\_proj}). Training is run for one epoch using an adapter trained jointly on the eleven dialects for which training data are available. This also supports transfer across varieties and enables zero-shot evaluation for Libyan and Sudanese, for which no training data are available. Zero-shot varieties are reached at inference only, through the natural-language dialect label and the model's pretraining exposure to related varieties. Appendix~\ref{app:hyper}, Table~\ref{tab:hyper} lists the configuration.

\subsection{Context-aware inference}
At test time, conversations are translated sequentially in dialogue order. Each Arabic translation the model produces for turn~$t$ is inserted into the context of turn~$t+1$. The dialogue context is therefore self-generated without using target-side information from the test set.  This prevents leakage but could lead to error propagation since a mistranslated turn remains visible for up to four subsequent turns. Decoding is greedy (\texttt{num\_beams}${=}1$) with a repetition penalty of 1.05 and a cap of 200 new tokens. For throughput, at each turn position prompts from 16 conversations are batched with left padding and generated in parallel. All predictions are cached to persistent storage per country and the pipeline can be restarted after interruptions. Figure~\ref{fig:pipeline} summarizes our system.

 \section{Experimental Setup}
\label{sec:setup}

\subsection{Data preparation}

We fold the dev split into training and create one supervised example for each dialogue turn. Each example contains the metadata, the preceding turns with their English sources and gold Arabic translations, and the gold Arabic translation of the current turn. This yields 78{,}730 examples (66{,}480 train + 12{,}250 dev). Because the official metric macro-averages over countries, we cap each country at 2{,}500 randomly
sampled examples. The cap applies to all eleven varieties with training
data and acts as a uniform downsample that equalises per-country counts (2{,}500 each, 27{,}500 in total) (Table~\ref{tab:data}).

\subsection{Preprocessing and training}

Prompts are limited to four preceding turns and truncated to 896 tokens, with older context removed first. We apply no orthographic normalization, dediacritization or transliteration because the Alexandria references preserve the original dialectal writing, including code-switched Latin-script tokens. We train for one epoch with a learning rate of $2\mathrm{e}{-4}$, cosine scheduling, 20 warmup steps, an effective batch size of 16, 1{,}719 optimizer steps, paged 8-bit AdamW and bf16 computation. Training uses one NVIDIA A100-SXM4-80GB GPU and takes approximately one hour. Decoding the private test set takes approximately 1.5 hours.

\subsection{Submission and evaluation}

We submit predictions for every private-test turn in both tracks using the official JSON format. The files are validated to ensure complete coverage and no missing, duplicate or empty predictions before submission. The primary metric is country-macro-averaged spBLEU, with chrF++ reported as a secondary metric.

\section{Results}
\label{sec:results}

\subsection{Official results}
Table~\ref{tab:leaderboard} presents the official results. Our system achieves 20.36 spBLEU and 36.64 chrF++, ranking sixth in both the constrained and unconstrained tracks. The same predictions were evaluated in both tracks.
Compared with the leading systems, we retain 80.7\% of the chrF++ score but only 66.9\% of the spBLEU score. Against the best unconstrained system, the figures are 76.1\% and 60.8\%. The chrF++/spBLEU ratio in Table~\ref{tab:leaderboard} increases monotonically with rank in both tracks, suggesting outputs that remain similar at the character level but differ in inflection, clitic attachment, spelling or lexical choice. 

\begin{table}[h]
\centering
\small
\setlength{\tabcolsep}{4pt}
\begin{tabular}{clrrr}
\toprule
\textbf{\#} & \textbf{Team} & \textbf{spBLEU} & \textbf{chrF++} & \textbf{Ratio} \\
\midrule
\multicolumn{5}{l}{\textit{Constrained track}} \\
1 & Thakaa          & 30.42 & 45.42 & 1.49 \\
2 & LahjaMT         & 28.30 & 43.81 & 1.55 \\
3 & FCDS            & 26.48 & 42.30 & 1.60 \\
4 & Rosetta         & 26.10 & 41.79 & 1.60 \\
5 & Namaa Community & 23.26 & 39.03 & 1.68 \\
6 & \textbf{Hasan\_kh (ours)} & \textbf{20.36} & \textbf{36.64} & \textbf{1.80} \\
\midrule
\multicolumn{5}{l}{\textit{Unconstrained track}} \\
1 & Thakaa          & 33.49 & 48.13 & 1.44 \\
2 & cepin           & 32.50 & 47.05 & 1.45 \\
3 & LahjaMT         & 28.54 & 44.02 & 1.54 \\
4 & Namaa Community & 27.41 & 42.58 & 1.55 \\
5 & Rosetta         & 25.09 & 41.02 & 1.63 \\
6 & \textbf{Hasan\_kh (ours)} & \textbf{20.36} & \textbf{36.64} & \textbf{1.80} \\
\bottomrule
\end{tabular}
\caption{Official Subtask~1 results.}
\label{tab:leaderboard}
\end{table}
\subsection{Post-submission analysis}
\label{sec:postsub}

These experiments were conducted post submission using matched public-test references and a fixed ablation set.

\paragraph{Code-switching.}
The system learns to code-switch in the varieties where the references use Latin-script material most frequently, especially Tunisian and Moroccan Arabic. However, its code-switching density is lower than the references in the main code-switching varieties, reaching 35--58\% of the reference token rate (0.58 for Tunisian, 0.58 for Moroccan, 0.48 for Lebanese, 0.35 for Egyptian and 0.15 for Mauritanian; Table~\ref{tab:codeswitch}).  We report percentage of turns containing at least one Latin-script token, percentage of all tokens in Latin script (token rates are Latin-script tokens divided by all whitespace-delimited tokens on that side's text), and the ratio of system to reference token rates. 
\begin{table}[h]
\centering\small
\setlength{\tabcolsep}{4pt}
\begin{tabular}{lrrrrr}
\toprule
& \multicolumn{2}{c}{\textbf{Turns (\%)}}
& \multicolumn{2}{c}{\textbf{Tokens (\%)}} & \\
\cmidrule(lr){2-3}\cmidrule(lr){4-5}
\textbf{Country} & \textbf{Ref} & \textbf{Sys} & \textbf{Ref} & \textbf{Sys}
& \textbf{Tok ratio} \\
\midrule
TN & 47.5 & 32.3 & 6.88 & 3.97 & 0.58 \\
MA & 29.5 & 18.9 & 4.02 & 2.34 & 0.58 \\
LB & 21.2 & 10.4 & 2.32 & 1.12 & 0.48 \\
EG & 10.2 &  4.3 & 0.94 & 0.33 & 0.35 \\
MR &  7.0 &  1.3 & 0.74 & 0.11 & 0.15 \\
SD &  4.0 &  1.5 & 0.46 & 0.14 & 0.30 \\
JO &  4.1 &  3.4 & 0.86 & 0.32 & 0.38 \\
SA &  2.2 &  1.3 & 0.23 & 0.11 & 0.47 \\
OM &  2.1 &  1.2 & 0.23 & 0.10 & 0.42 \\
PS &  1.6 &  0.9 & 0.24 & 0.08 & 0.35 \\
YE &  1.0 &  0.5 & 0.10 & 0.05 & 0.53 \\
LY &  1.4 &  1.7 & 0.11 & 0.14 & 1.27 \\
SY &  0.8 &  1.3 & 0.08 & 0.13 & 1.63 \\
\bottomrule
\end{tabular}
\caption{Latin-script code-switching on the matched public-test pair.}
\label{tab:codeswitch}

\end{table}

Replacing omitted loanwords with Arabic paraphrases reduces $n$-gram overlap and likely contributes to the lower spBLEU scores. In the two zero-shot varieties the direction reverses. The system over-produces Latin-script tokens relative to the references (1.27 for Libyan, 1.63 for Syrian). Under-production is therefore specific to the dialects whose references code-switch heavily, while the consistent relationship between the dialect label and code-switching density shows that the natural-language dialect label affects register-level behaviour, not only vocabulary.

\paragraph{Context ablation.}
We compared self-generated history, gold Arabic history and empty history on 200 conversations per country (Table~\ref{tab:ablation}). 

\begin{table}[h]
\centering\footnotesize
\begin{tabular}{lrr}
\toprule
\textbf{Context condition} & \textbf{spBLEU} & \textbf{chrF++} \\
\midrule
B: gold Arabic history      & 21.62 & 37.66 \\
C: empty history            & 21.41 & 37.23 \\
A: self-generated (submitted) & 20.56 & 36.73 \\
\bottomrule
\end{tabular}
\caption{Context ablation on 200 conversations per country (8{,}000 turns), scored against public-test references.}
\label{tab:ablation}
\end{table}
Gold history achieves 21.62 macro spBLEU and 37.66 chrF++, while empty history achieves 21.41 and 37.23. The submitted self-generated condition performs worst, with 20.56 spBLEU and 36.73 chrF++. This shows that the model gains little from dialogue history and that errors in its previous translations can outweigh the benefits of context. The effect is consistent and empty history matches or outperforms the submitted self-generated condition in all thirteen varieties (Table~\ref{tab:ablation_country}). 

\begin{table}[h]
\centering\small
\setlength{\tabcolsep}{4pt}
\begin{tabular}{lrrrrr}
\toprule
\textbf{Country} & \textbf{A: self} & \textbf{B: gold} & \textbf{C: empty} & \textbf{B$-$A} & \textbf{C$-$A} \\
\midrule
SY & 26.88 & 27.48 & 28.08 & +0.60 & +1.20 \\
JO & 25.95 & 27.91 & 27.07 & +1.96 & +1.12 \\
TN & 25.29 & 26.38 & 26.32 & +1.09 & +1.03 \\
SA & 24.78 & 26.84 & 26.02 & +2.06 & +1.24 \\
EG & 23.80 & 24.81 & 24.43 & +1.01 & +0.63 \\
PS & 21.02 & 22.74 & 22.60 & +1.72 & +1.58 \\
LB & 20.99 & 21.46 & 21.68 & +0.47 & +0.69 \\
OM & 19.80 & 20.56 & 21.05 & +0.76 & +1.25 \\
MA & 19.30 & 19.85 & 20.00 & +0.55 & +0.70 \\
YE & 18.88 & 19.72 & 19.36 & +0.84 & +0.48 \\
SD & 16.13 & 17.27 & 16.91 & +1.14 & +0.78 \\
LY & 14.74 & 15.45 & 14.85 & +0.71 & +0.11 \\
MR &  9.75 & 10.55 &  9.94 & +0.80 & +0.19 \\
\midrule
\textbf{Macro} & \textbf{20.56} & \textbf{21.62} & \textbf{21.41} & \textbf{+1.06} & \textbf{+0.85} \\
\bottomrule
\end{tabular}
\caption{Per-country context ablation (macro spBLEU).}
\label{tab:ablation_country}
\end{table}

\paragraph{Orthographic normalization.}
Normalizing system outputs and references with five rulesrules---given in Appendix~\ref{app:norm}---increases macro spBLEU from 20.80 to 22.36 and chrF++ from 37.01 to 38.28. The largest spBLEU improvements occur for Saudi (+2.68), Omani (+2.56),
Sudanese (+2.11) and Egyptian (+2.08) Arabic, and the smallest for
Moroccan (+0.64) and Syrian (+0.73) (Table~\ref{tab:normalization}, and per-country gains in Appendix~\ref{app:percountry}, Table~\ref{tab:percountry}).
However, normalizing system outputs only, while leaving references unchanged, reduces macro spBLEU from 20.80 to 15.49 ($-5.31$). This suggests that the model already matches the spelling style of the corpus. Changing only the model outputs breaks matches with the references. Normalization is therefore only useful for analysis, when both the model outputs and the references are normalized in the same way. It should not be applied to the submitted outputs as it would lower the score. 

\begin{table}[h]
\centering\small
\begin{tabular}{lrrr}
\toprule
 & \textbf{Raw} & \textbf{Normalised} & \textbf{$\Delta$} \\
\midrule
spBLEU (macro) & 20.80 & 22.36 & +1.56 \\
chrF++ (macro) & 37.01 & 38.28 & +1.27 \\
\bottomrule
\end{tabular}
\caption{Effect of orthographic normalzation on the public-test evaluation. }
\label{tab:normalization}
\end{table}

\paragraph{MSA fallback.}
The system does not show a strong general tendency to fall back to MSA. Its MSA function-word rates are close to the reference rates, with only small positive residuals for Sudanese (+0.34), Tunisian (+0.35) and Mauritanian (+0.18) (Table~\ref{tab:msa}). The apparent MSA signal in the raw system outputs therefore largely reflects the reference texts and not systematic model behavior.

\begin{table}[h]
\centering\footnotesize
\begin{tabular}{lc}
\toprule
\textbf{Country} & \textbf{MSA $\Delta$ (system $-$ reference) } \\
\midrule
TN & $+0.35$ \\ SD & $+0.34$ \\ MR & $+0.18$ \\ EG & $+0.16$ \\
SY & $+0.09$ \\ LY & $+0.09$ \\ JO & $+0.08$ \\ OM & $+0.07$ \\
SA & $+0.04$ \\ LB & $\phantom{+}0.00$ \\ MA & $-0.10$ \\
PS & $-0.01$ \\ YE & $+0.06$ \\
\bottomrule
\end{tabular}
\caption{MSA function-word rate difference.}
\label{tab:msa}
\end{table}

\paragraph{Manual error analysis.}
We annotated 60 private-test turns, approximately five per country, using categories adapted from the shared task's LQM-inspired typology. Twenty-three turns were error-free (Appendix~\ref{app:annotation}, Table~\ref{tab:annotation}). Content errors were most common (19), followed by dialect--MSA confusion (11), which was concentrated in Libyan (3), Sudanese (3) and Mauritanian (2). Morphological agreement errors occurred in 8 turns across eight countries, while orthographic errors occurred in 2. Six further turns had understandable content but sounded non-native in register. The LLM annotator identified the content and morphology errors but missed these foreign-register cases, showing the value of human evaluation for dialectal naturalness.

\section{Conclusion}
\label{sec:conclusion}
We present a compact, dialect-focused 3B model fine-tuned with QLoRA on turn-level metadata-rich prompts. This provides a practical constrained track system for context-aware dialectal Arabic dialogue translation. The system trains in about one GPU-hour, feeds its own outputs back as context without any gold leakage at inference, and covers thirteen dialects. It achieves 20.36 spBLEU / 36.64 chrF++ on both tracks. Future work covers longer training with a larger LoRA rank and beam-search decoding.

\bibliography{custom}
%\newpage
\appendix
%\newpage
\section{Prompt Template}
\label{app:prompt}

Prompts are generated following the template below. Angle brackets mark fields filled from the dataset. In the actual prompt, \texttt{<arabic\_translation\_i>} contains the gold Arabic during training and the model's own previous output at inference.

\begin{quote}\small\ttfamily
\noindent Translate ONLY the last English turn into <dialect\_name>. Preserve the meaning, tone, register, persona and gender agreement of the speaker. Output the translation and nothing else.\\[0.5em]
\#\#\# Metadata\\
Country: <country\_name> (<country\_code>)\\
Dialect: <dialect\_name>\\
Domain: <domain>\\
Participants: <persona\_a>; <persona\_b>\\
Speaker: <speaker\_role> (<gender\_direction>)\\[0.5em]
\#\#\# Conversation so far\\
Turn <i> [<speaker\_i>] EN: <english\_turn\_i>\\
Turn <i> [<speaker\_i>] AR: <arabic\_translation\_i>\\
\textnormal{\rmfamily (last four turns only)}\\[0.5em]
\#\#\# Turn to translate\\
EN: <english\_turn\_t>\\
AR:
\end{quote}

\noindent The model generates after the final marker. At inference, generation stops at the end-of-sequence token or after 200 new tokens.

\newpage

\section{Full Configuration}
\label{app:hyper}

\begin{table}[htbp!]
\centering
\small
\setlength{\tabcolsep}{4pt}
\begin{tabular}{p{0.30\columnwidth}p{0.62\columnwidth}}
\toprule
\textbf{Component} & \textbf{Setting} \\
\midrule
Base model & UBC-NLP/NileChat-3B-Base ($\approx$3B params; limit 5B) \\
Quantisation & 4-bit NF4, double quantisation, bf16 compute (bitsandbytes) \\
LoRA & $r=16$, $\alpha=32$, dropout $0.05$; targets \texttt{q/k/v/o\_proj}, \texttt{gate/up/down\_proj} \\
Epochs & 1 \\
Learning rate & $2\mathrm{e}{-4}$, cosine schedule, 20 warmup steps \\
Batch size & 8 per device $\times$ 2 grad.\ accumulation (effective 16); 1{,}719 steps \\
Max seq.\ length & 896 tokens (left truncation) \\
Context window & last 4 turns \\
Optimizer & paged AdamW 8-bit; bf16; gradient checkpointing disabled \\
Training data & 27{,}500 examples (per-country cap 2{,}500; train + dev) \\
Decoding & greedy (\texttt{num\_beams}=1), repetition penalty 1.05, max 200 new tokens \\
Inference batching & 16 prompts, left padding \\
Hardware & 1 $\times$ NVIDIA A100-SXM4-80GB (Colab); $\approx$1\,h train, $\approx$1.5\,h inference \\
Software & transformers 5.14.1, trl 1.9.0, peft 0.19.1, torch 2.11 (CUDA 12.8) \\
\bottomrule
\end{tabular}
\caption{System configuration}
\label{tab:hyper}
\end{table}

\section{Orthographic Normalization Rules}
\label{app:norm}
The following
mappings are applied to both hypotheses and references before rescoring. Codepoints are given alongside the glyphs to avoid ambiguity.

\begin{itemize}
\itemsep0em
\item Alef variants \foreignlanguage{arabic}{أ} (U+0623), \foreignlanguage{arabic}{إ} (U+0625) and \foreignlanguage{arabic}{آ} (U+0622)
      $\rightarrow$ bare alef \foreignlanguage{arabic}{ا} (U+0627)
\item Alef maqsura \foreignlanguage{arabic}{ى} (U+0649) $\rightarrow$ ya \foreignlanguage{arabic}{ي} (U+064A)
\item Ta marbuta \foreignlanguage{arabic}{ة} (U+0629) $\rightarrow$ ha \foreignlanguage{arabic}{ه} (U+0647)
\item Hamza carriers: waw with hamza \foreignlanguage{arabic}{ؤ} (U+0624) $\rightarrow$ waw
      \foreignlanguage{arabic}{و} (U+0648); ya with hamza \foreignlanguage{arabic}{ئ} (U+0626) $\rightarrow$ ya
      \foreignlanguage{arabic}{ي} (U+064A)
\item Deletion of short-vowel and gemination diacritics
      \foreignlanguage{arabic}{ـَ ـُ ـِ ـً ـٌ ـٍ ـّ ـْ} (U+064B--U+0652), superscript alef
      \foreignlanguage{arabic}{ـٰ} (U+0670), and tatweel \foreignlanguage{arabic}{ـ} (U+0640)
\item Collapse of repeated whitespace
\end{itemize}

\newpage
\section{Per-Country Results}
\label{app:percountry}

\begin{table}[h]
\centering\footnotesize
\setlength{\tabcolsep}{3.5pt}
\begin{tabular}{lrrrrr}
\toprule
& \multicolumn{2}{c}{\textbf{Public test}} & \textbf{Norm.} \\
\cmidrule(lr){2-3}\cmidrule(lr){4-4}
\textbf{Ctry} & \textbf{spBLEU} & \textbf{chrF++} & \textbf{$\Delta$} \\
\midrule
SY & 26.31 & 42.88 & $+0.73$  \\
JO & 26.00 & 42.14 & $+1.33$  \\
TN & 25.11 & 39.37 & $+0.79$ \\
SA & 24.74 & 40.92 & $+2.68$  \\
EG & 23.41 & 38.33 & $+2.08$  \\
LB & 21.80 & 37.87 & $+1.65$ \\
PS & 21.27 & 37.63 & $+1.58$ \\
OM & 20.14 & 36.75 & $+2.56$ \\
SD$^{*}$ & 18.69 & 34.89 & $+2.11$ \\
YE & 18.48 & 35.16 & $+1.29$ \\
MA & 18.02 & 34.71 & $+0.64$ \\
LY$^{*}$ & 16.38 & 33.37 & $+1.19$\\
MR & 10.10 & 27.15 & $+1.57$\\
\midrule
Macro & 20.80 & 37.01 & $+1.56$\\
\bottomrule
\end{tabular}
\caption{Per-country results on the public test split. \textbf{Norm.} is the
spBLEU gain from orthographic normalization. $^{*}$zero-shot varieties.}
\label{tab:percountry}
\end{table}

\section{Annotation Protocol}
\label{app:annotation}

We sampled 60 turns from our private-test predictions, stratified to give
approximately five per country. Each turn was presented with its English source,
the preceding dialogue context, the country and dialect label, and the
speaker--addressee gender direction, but without a reference translation, since
references for the private test set are not released.

Labels were assigned independently by one native-speaker annotator and by an LLM
prompted with the same category definitions, and were then reconciled by the
human annotator. Categories were adapted from the LQM-inspired typology of
Subtask~3 \citep{magdy-etal-2026-lqm}: content or meaning error, dialect--MSA
confusion, morphological agreement error, orthographic error, and register or
persona mismatch. A sixth category, \textit{unnatural but content-acceptable},
was added by the human annotator during labelling for turns whose content was
recoverable but whose register sounded non-native.

\begin{table}[h]
\centering\small
\begin{tabular}{lr}
\toprule
\textbf{Category} & \textbf{Turns} \\
\midrule
Error-free                        & 23 \\
Content / meaning error           & 19 \\
Dialect--MSA confusion            & 11 \\
Morphological agreement           &  8 \\
Unnatural, content acceptable     &  6 \\
Orthographic error                &  2 \\
Register / persona mismatch       &  1 \\
\bottomrule
\end{tabular}
\caption{Annotation counts over 60 sampled turns.}
\label{tab:annotation}
\end{table}

\end{document}